\documentclass[10pt,twocolumn,letterpaper]{article}
\usepackage[pagenumbers]{cvpr} 

\definecolor{cvprblue}{rgb}{0.21,0.49,0.74}
\usepackage[table]{xcolor}
\usepackage{pifont}
\usepackage[most]{tcolorbox} 
\usepackage{xcolor} 
\usepackage{lipsum} 
\usepackage{cuted} 
\usepackage{threeparttable}
\usepackage{hyperref}

\newtcolorbox{theorembox}{
  colback=blue!20!cyan!10!white!50, 
  colframe=gray!100!black,  
  arc=1mm, 
  boxrule=1pt,  
  left=5pt,right=5pt,top=5pt,bottom=5pt,  
  width=0.99\linewidth 
}

\newenvironment{mythm}[2][]{
  \begin{theorembox}
  \textbf{Proposition: #2}#1. \itshape
}{
  \end{theorembox}
}

\newenvironment{myfinding}[2][]{
  \begin{theorembox}
  \textbf{Observation: #2}#1. \itshape
}{
  \end{theorembox}
}

\title{Neural Collapse in Test-Time Adaptation}

\author{
{Xiao Chen}$^1$, 
{Zhongjing Du}$^{2}$, 
{Jiazhen Huang}$^{1}$,
{Xu Jiang}$^2$,  \\
{Li Lu}$^3$, 
{Jingyan Jiang}$^4$\footnotemark[1], 
{Zhi Wang}$^1$\footnotemark[1]
\\
$^1$ Shenzhen International Graduate School, Tsinghua University
\\
$^2$ Peking University
$^3$ Sichuan University
$^4$ Shenzhen Technology University \\
\small{
\texttt{chen-x25@mails.tsinghua.edu.cn}, \texttt{jiangjingyan@sztu.edu.cn}, \texttt{wangzhi@sz.tsinghua.edu.cn}
}}

\begin{document}
\maketitle

\renewcommand{\thefootnote}{\fnsymbol{footnote}}
\footnotetext[1]{Co-corresponding authors}
\renewcommand{\thefootnote}{\arabic{footnote}}

\begin{abstract}
  Test-Time Adaptation (TTA) enhances model robustness to out-of-distribution (OOD) data by updating the model online during inference, yet existing methods lack theoretical insights into the fundamental causes of performance degradation under domain shifts. Recently, Neural Collapse (NC) has been proposed as an emergent geometric property of deep neural networks (DNNs), providing valuable insights for TTA. In this work, we extend NC to the sample-wise level and discover a novel phenomenon termed Sample-wise Alignment Collapse (NC3+), demonstrating that a sample's feature embedding, obtained by a trained model, aligns closely with the corresponding classifier weight. Building on NC3+, we identify that the performance degradation stems from sample-wise misalignment in adaptation which exacerbates under larger distribution shifts. This indicates the necessity of realigning the feature embeddings with their corresponding classifier weights. However, the misalignment makes pseudo-labels unreliable under domain shifts. To address this challenge, we propose NCTTA, a novel feature-classifier alignment method with hybrid targets to mitigate the impact of unreliable pseudo-labels, which blends geometric proximity with predictive confidence. Extensive experiments demonstrate the effectiveness of NCTTA in enhancing robustness to domain shifts. For example, NCTTA outperforms Tent by 14.52\% on ImageNet-C. Project page is publicly available at \url{https://github.com/Cevaaa/NCTTA}.
\end{abstract}

\begin{figure}
    \vspace{-10pt}
    \centering
    \includegraphics[width=0.9\linewidth]{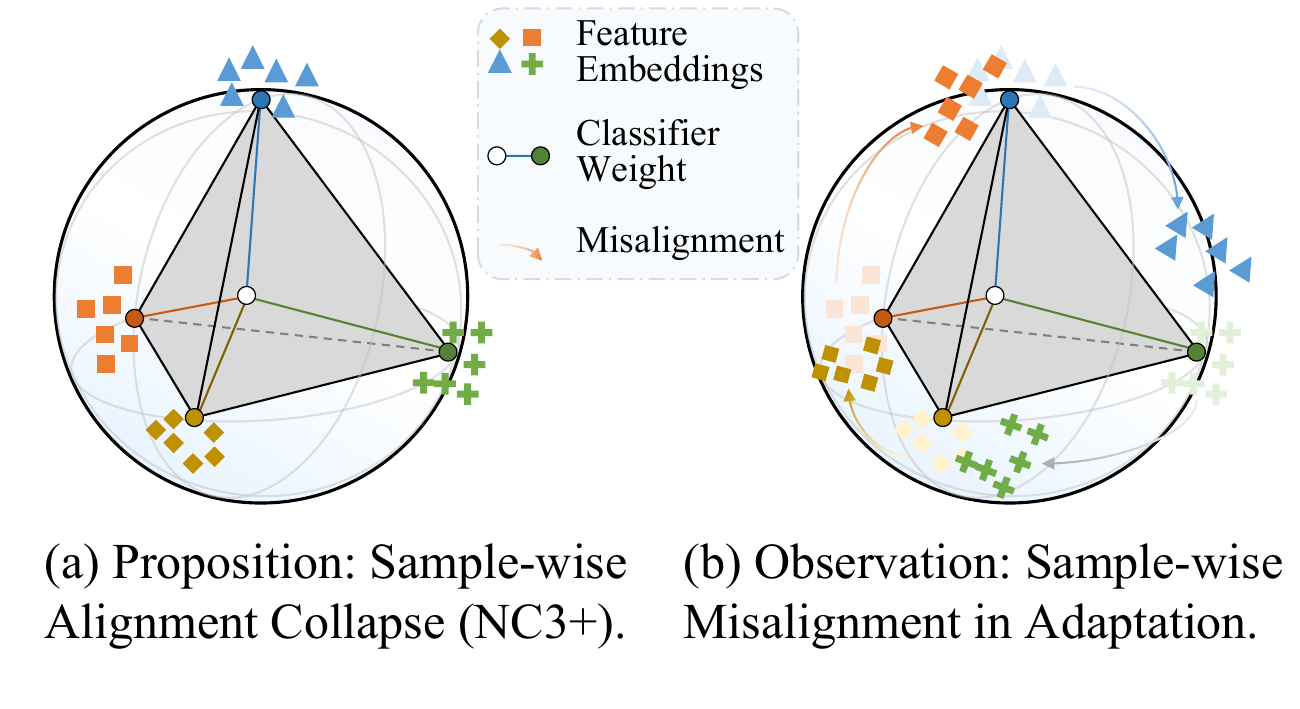}
    \captionsetup{width=0.99\linewidth}
    \vspace{-15pt}
    \caption{
        \textbf{Overview of Main Contributions.} 
        (a) NC3+ highlights the convergence of sample feature embeddings with their corresponding classifier weights. 
        (b) Sample feature embeddings deviate from the ground-truth classifier weights, leading to performance degradation.
    }
    \label{intro}
    \vspace{-18pt}
\end{figure}
\par
\noindent

\section{Introduction}
\label{sec:intro}
Deep neural networks (DNNs) \cite{DNN} have achieved remarkable success across various tasks but suffer significant performance degradation under out-of-distribution (OOD) data. Traditional approaches like Domain Adaptation (DA) \cite{DG_survey, UDA_survey} aim to address this issue but often involve complex training procedures and assume access to test data, which is rarely feasible in real-world. Test-Time Adaptation (TTA) offers a more practical solution by enabling models to adapt online during inference using only a mini-batch of test data, without altering the training process \cite{TTA_survey, fan2025moetta}. This lightweight and efficient paradigm improves robustness to domain shifts, facilitating reliable deployment in dynamic and unpredictable environments.

Existing TTA methods can be broadly categorized into four main strategies: prototype-based methods \cite{shot, t3a}, consistency regularization methods~\cite{Memo, cotta}, normalization layer-based methods~\cite{bn_adapt, note, SAR, jiang2025feature}, and entropy minimization methods \cite{Tent, EATA, DEYO, cliff}. Although these approaches have shown promising results through algorithmic optimization during inference, they generally lack a theoretical exploration of the fundamental causes of model degradation when facing domain shifts.

Recently, the phenomenon known as \textit{Neural Collapse} (NC)~\cite{neural_collapse, ncseg, ncvlm, ncooddetector} has emerged as a significant characterization of the geometric properties of DNNs during the Terminal Phase of Training (TPT). During the TPT, after achieving zero classification error, DNNs asymptotically drive the training loss toward zero. Unlike previous TTA methods, this work bridges the theory of NC with TTA by leveraging the well-defined geometric structures of a trained DNN formalized by NC properties. This connection offers novel theoretical insights into TTA, grounding it in the structured behavior of DNNs. Specifically, during the TPT, NC consistently appears in the penultimate layer and its linear classifier~\cite{minor_collapse}, exhibiting a highly structured and intriguing behavior defined by four properties:

\begin{itemize} 
\item \textbf{Variability Collapse (NC1):} Sample feature embeddings within the same class collapse to their corresponding class means, reducing intra-class variance. 
\item \textbf{Convergence to a Simplex ETF (NC2):} Class means converge to the vertices of a simplex equiangular tight frame (ETF), forming a maximally separated structure. 
\item \textbf{Convergence to Self-Duality (NC3):} Class means and corresponding classifier weights will converge to each other during the TPT.
\item \textbf{Simplification to Nearest Class-Center (NC4):} The classification function effectively reduces to choosing the nearest class mean in the feature space. 
\end{itemize}

While NC unveils elegant geometric patterns in DNNs during the TPT, its analysis predominantly focuses on a \textit{class-wise} perspective. However, this assumption breaks down in TTA scenarios, where models are adapted using only mini-batches of unlabeled test samples. The absence of labeled data precludes the computation of \textit{class means}, which are indispensable in NC properties, thereby rendering class-wise analyses inapplicable in TTA. This challenge motivates us to explore whether similar collapse phenomena emerge at the sample-wise level. As illustrated in Figure~\ref{intro}(a), our analysis extends the NC framework by identifying an additional property, NC3+, that holds at the sample-wise granularity:

\begin{itemize} 
\item \textbf{Sample-wise Alignment Collapse (NC3+):} Each sample feature embedding and the corresponding classifier weight will converge to each other during the TPT. 
\end{itemize}

Building on this observation, we further analyze the performance degradation under domain shifts during test time. As illustrated in Figure~\ref{intro}(b), for OOD data, the feature embedding of a test sample tends to deviate from its corresponding classifier weight and drift toward that of an incorrect one, leading to misclassification. This misalignment intensifies as the domain shift becomes more severe, highlighting a key challenge in TTA:

\noindent\textit{Misalignment between feature embeddings and classifier weights is the key challenge under domain shifts, which underscores the necessity of feature-classifier alignment.}

To mitigate this effect, we propose a novel \textbf{NC}-guided \textbf{TTA} (NCTTA) method to promote feature-classifier alignment. The sample-wise misalignment implies the unreliability of pseudo-labels. Therefore, instead of simply aligning the feature embedding with the classifier weight of its pseudo-label, NCTTA blends geometric proximity with predictive confidence to obtain hybrid targets for realignment. This strategy substantially improves robustness under domain shifts by alleviating the pseudo-label unreliability caused by Sample-wise Misalignment in Adaptation. Our main contributions are as follows: 
\begin{itemize}
\item NC theory is, for the first time, extended to the sample-wise level through Sample-wise Alignment Collapse (NC3+), validated theoretically and empirically.
\item Comprehensive empirical evidence is provided to show that the performance degradation of pre-trained models on OOD data originates from sample-wise misalignment.
\item A novel NC-guided TTA method is proposed to effectively promote alignment between sample feature embeddings and their corresponding classifier weights.
\item Extensive experiments demonstrate the effectiveness of NCTTA. For instance, it achieves 78.30\% average accuracy on CIFAR-10-C under severe corruption and 66.61\% on ImageNet-C, outperforming prior methods.
\end{itemize}

\section{Related Work}

\label{sec:rw}
\subsection{Test-Time Adaptation}  

\label{subsec:rw-tta}  
TTA addresses performance degradation on OOD data by fine-tuning a pretrained model during inference using a mini-batch in an online manner~\cite{shift_1,shift_2}. Existing TTA methods can be categorized into four types: prototype-based methods (e.g., SHOT~\cite{shot}, T3A~\cite{t3a}) refine decision boundaries using class prototypes; consistency regularization methods (e.g., MEMO~\cite{Memo}, CoTTA~\cite{cotta}) stabilize predictions through constraints across augmented views or over time; normalization-based methods (e.g., NOTE~\cite{note}, SAR~\cite{SAR}) explore the role of normalization layers in TTA and carefully design their adaptation; and entropy minimization methods (e.g., Tent~\cite{Tent}, EATA~\cite{EATA}, DeYO~\cite{DEYO}) enhance reliability by minimizing prediction entropy with confident samples. Despite their empirical effectiveness, these methods lack theoretical understanding of model behavior under domain shifts. To address this, we provide a principled analysis of performance degradation during TTA and propose a theory-driven method that aligns test-time feature embeddings with their corresponding classifier weights, improving robustness to OOD scenarios.

\subsection{Neural Collapse}  

\label{sebsec:rw-nc}  
NC describes a set of geometric properties that emerge in DNNs during the TPT~\cite{neural_collapse}. These properties include the collapse of within-class variability (NC1), the convergence of class means to a Simplex Equiangular Tight Frame (NC2), the alignment of classifier weights with class means (NC3), and classification via the nearest class mean (NC4). These phenomena have been consistently observed across diverse datasets and architectures under both cross-entropy and MSE losses~\cite{minor_collapse}, and are often interpreted as optimal structures arising from simplified learning objectives~\cite{ncseg, ncooddetector, ncvlm}.  However, existing works on NC~\cite{zhu2021geometric, lu2022neural, fang2021layer, weinan2022emergence, zhou2022all} have primarily focused on class-wise geometric structures between class centers and classifier weights observed in the TPT, relying on class labels and full supervision to define its geometric properties, whereas TTA operates on unlabeled target data with incremental updates using mini-batches, making NC's class-wise assumptions inapplicable. To overcome this limitation, NC is extended to a sample-wise perspective through the introduction of NC3+, as detailed in Section~\ref{subsec:rnc-enc}. Building on the insights provided by NC3+, the causes of misclassification on OOD data are analyzed in Section~\ref{subsec:tta-degradation}. To address these challenges, NCTTA is proposed, a feature-classifier alignment method, leveraging the principles of NC to enhance TTA performance under domain shifts.


\section{Rethinking Neural Collapse}

\label{sec:rnc}
\subsection{Preliminaries}

\label{subsec:rnc-pre}
In this section, we establish the notations used throughout the paper. Let the training and test datasets be denoted as $
D_{\text{train}} = \{(x_i, y_i)\}_{i=1}^{N_\text{train}} \sim \mathcal{P}_{\text{train}}$ and $D_{\text{test}} = \{x_i\}_{i=1}^{N_\text{test}} \sim \mathcal{P}_{\text{test}}
$, where $N_\text{train}$ and $N_\text{test}$ represent the number of samples in the training and test sets, respectively. Under the TTA setting, there exists a domain shifts such that $\mathcal{P}_{\text{train}} \neq \mathcal{P}_{\text{test}}$, and the $D_{\text{test}}$ is unlabeled. Moreover, during adaptation, only a mini-batch of test samples is accessible at each iteration.

We denote a DNN by $f_\theta(\cdot)$, which consists of a feature extractor $h_{\theta - \omega}(\cdot)$ and a classifier head $g_{\omega}(\cdot)$, i.e.,
$f_\theta(\cdot) = (g_{\omega} \circ h_{\theta - \omega})(\cdot)$,
where $\circ$ denotes function composition. Here, $\theta$ represents the full set of model parameters, and $\omega = [w_1, w_2, \dots, w_K] \in \mathbb{R}^{K \times L}$ denotes the classifier weights, with $K$ the number of classes and $L$ the feature embedding dimension. For simplicity, we omit the bias term in the classifier.

Given an input sample $x_i$, its feature embedding is computed as $\mathbf{h}_i = h_{\theta - \omega}(x_i) \in \mathbb{R}^{L}$. The corresponding logits are given by $
\mathbf{z}_i = g_{\omega}(\mathbf{h}_i) \in \mathbb{R}^{K} $.  Applying the softmax function $\sigma(\cdot)$ to $\mathbf{z}_i$ yields the predicted class probabilities $
p_i = \sigma(\mathbf{z}_i) \in \mathbb{R}^{K}$, where $ p_{ij} = \exp(\mathbf{z}_{ij})/\sum_{k=1}^{K} \exp(\mathbf{z}_{ik})$.

\subsection{Extension of Neural Collapse}

\label{subsec:rnc-enc}
This subsection analyzes the training phase of $f_\theta$, focusing on the NC3. The remaining NC properties are detailed in Appendix A.1. NC3 describes the \emph{self-duality} phenomenon, where class means from $h_{\theta - \omega}$ align with the classifier weights $\omega$ in $g_{\omega}$, up to a rescaling factor. Formally, the \emph{global mean} and \emph{class mean} for each class $c$ are defined as: 
$\mu_G = 1/|D_{\text{train}}| \sum_{x_i \in D_{\text{train}}} h_{\theta - \omega}(x_i)$,
$\mu_c = 1/|D_{\text{train}}^c| \sum_{x_i \in D_{\text{train}}^c} h_{\theta - \omega}(x_i)$, 
where $D_{\text{train}}^c = \{ (x_i, y_i) \in D_{\text{train}} \mid y_i = c \}$ and $| \cdot |$ denotes the cardinality of a set. Prior work~\cite{neural_collapse} quantify NC3 with the metric:
\begin{equation}
    \left\| \frac{\mathbf{M}}{\|\mathbf{M}\|_2} - \frac{\omega}{\|\omega\|_2} \right\|_2 \;\longrightarrow\; 0,
\end{equation}
where $\mathbf{M} = [\mu_0 - \mu_G, \mu_1 - \mu_G, \dots, \mu_{K-1} - \mu_G]$. 

NC3 captures the convergence of class means and classifier weights, revealing an elegant alignment between feature and parameter spaces. However, it is inherently class-centric, relying on class labels to compute $\mu_c$ and the full training set to estimate $\mu_G$. These assumptions are incompatible with TTA, where only unlabeled mini-batches are accessible during inference. To address this, we propose NC3+, a sample-wise extension of NC3 that examines the alignment between individual sample feature embeddings and classifier weights. Specifically, we define a Feature-Classifier Alignment (FCA) distance $d_{ij}$ of a test sample \(x_i\) as the normalized Euclidean distance between its feature embedding $\mathbf{h}_i = h_{\theta - \omega}(x_i)$ and classifier weight $w_j$ of \(j\)-th class. Formally, the FCA distance $d_{ij}$ is calculated as:
\begin{equation}
    d_{ij} = \left\| \frac{\mathbf{h}_i}{\|\mathbf{h}_i\|_2} - \frac{w_j}{\|w_j\|_2} \right\|_2, \quad j = 0, 1, \dots, K-1.
\end{equation}

To validate NC3+ during the TPT, we provide a theoretical proof (Appendix A.3) demonstrating that, for a labeled sample $(x_i, y_i) \in D_{\text{train}}$, with the cross-entropy loss $\mathcal{L}_{\text{CE}}(x_i, y_i) = -\log p_{y_i}$, the Ground-truth FCA (G-FCA) distance $d_{iy_i}$ decreases monotonically and converges to zero. In contrast, for $j \neq y_i$, $d_{ij}$ remains largely unaffected. We formally define NC3+ as:
\begin{mythm}{Sample-wise Alignment Collapse (NC3+)} During the TPT, the G-FCA distance \(d_{iy_i}\) between a sample's feature embedding and its corresponding classifier weight is expected to converge to zero, formulated as:
\begin{equation}
    \text{G-FCA:}\quad d_{iy_i} = \left\| \frac{\mathbf{h}_i}{\|\mathbf{h}_i\|_2} - \frac{w_{y_i}}{\|w_{y_i}\|_2} \right\|_2 \;\longrightarrow\; 0.
\end{equation}
\end{mythm}

\begin{figure}[htbp]
    \centering
    \vspace{-8pt}
    \includegraphics[width=0.9\linewidth]{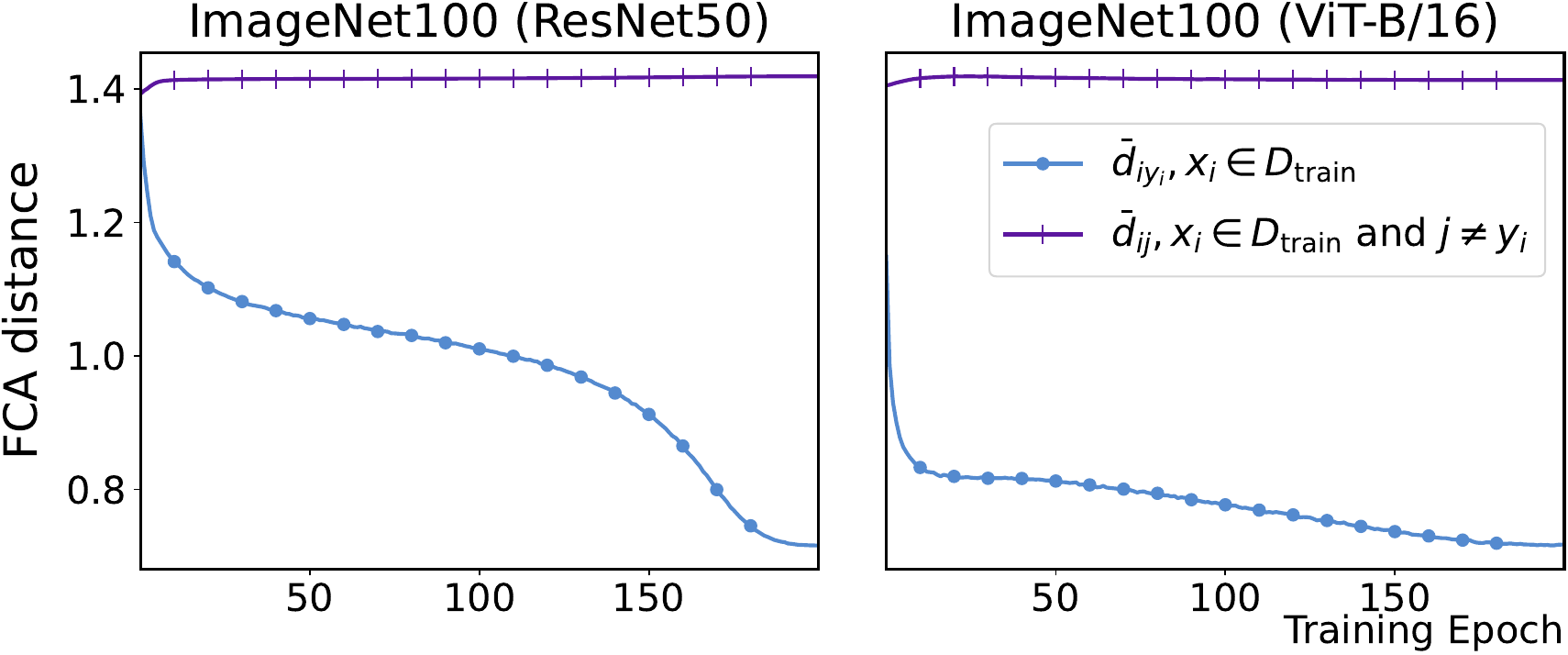}
    \vspace{-8pt}
    \caption{\textbf{Empirical validation of NC3+.} We evaluate NC3+ on ImageNet-100~\cite{imagenet100} using various backbones. The G-FCA distance $d_{iy_i}$ decreases throughout training, indicating sample-wise alignment collapse. More details are provided in Appendix A.2.}
    \label{fig:nc3plus}
    \vspace{-8pt}
\end{figure}

NC3+ encapsulates the phenomenon of sample-wise alignment collapse during the TPT, extending NC theory to settings where class-wise and global means are not feasible. We empirically validate it through controlled experiments. As shown in Figure~\ref{fig:nc3plus}, G-FCA distance $d_{iy_i}$ consistently decreases and asymptotically approaches zero during training, substantiating the presence of NC3+. \textit{This implies that well-trained DNNs generally satisfy the NC3+ property, extending the Neural Collapse phenomenon to settings where class-wise and global means are not feasible.}

\subsection{Explaining Performance Degradation in TTA}
\label{subsec:tta-degradation}
\begin{figure}[htbp]
    \centering
    \includegraphics[width=0.9\linewidth]{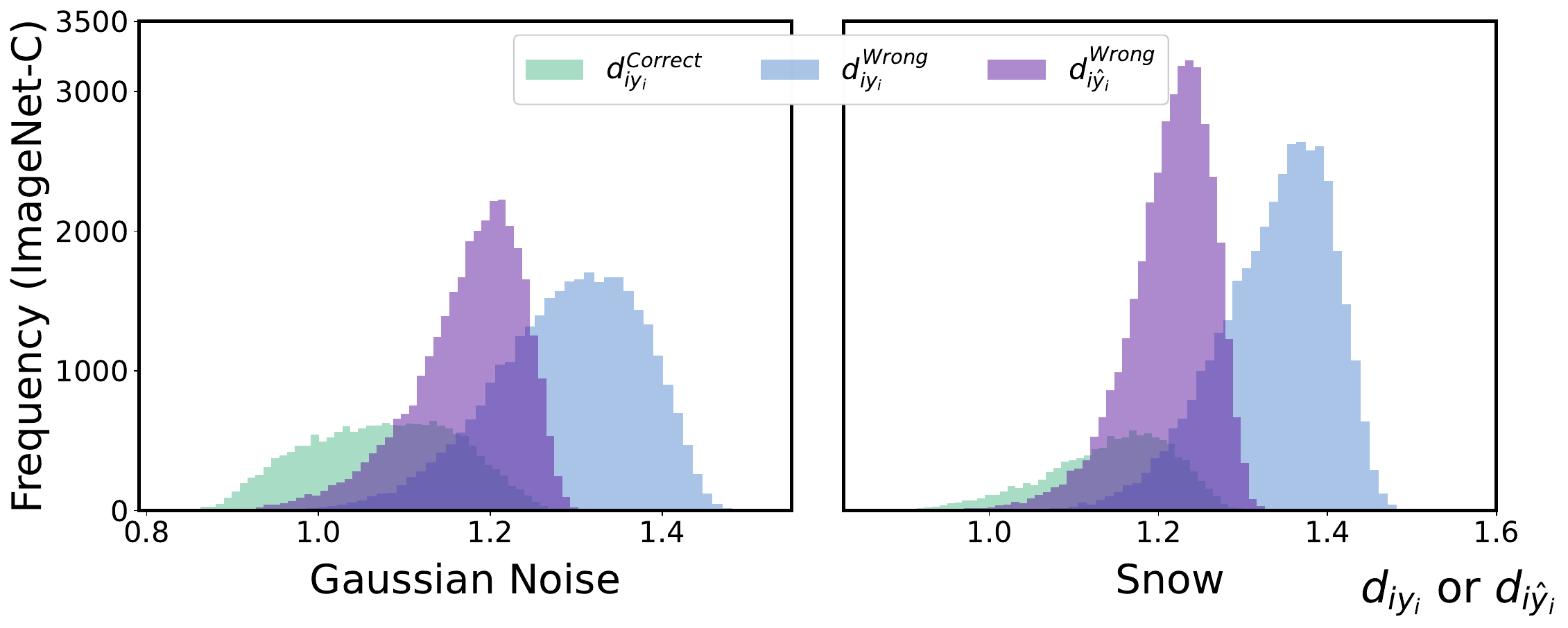}
    \vspace{-10pt}
    \caption{\textbf{Histograms of G-FCA and P-FCA distances.} We present the distributions of $d_{iy_i}^{\text{correct}}$, $d_{iy_i}^{\text{wrong}}$, and $d_{i\hat{y}_i}^{\text{wrong}}$ on correctly and incorrectly classified samples from ImageNet-C datasets under severity level 5 Gaussian noise and Snow corruption. The results reveal that NC3+ is violated on OOD data, leading to Sample-wise Misalignment in Adaptation.}
    \vspace{-10pt}
    \label{fig:histogram-tta}
\end{figure}

\begin{figure}[htbp]
    \centering  
    \includegraphics[width=0.9\linewidth]{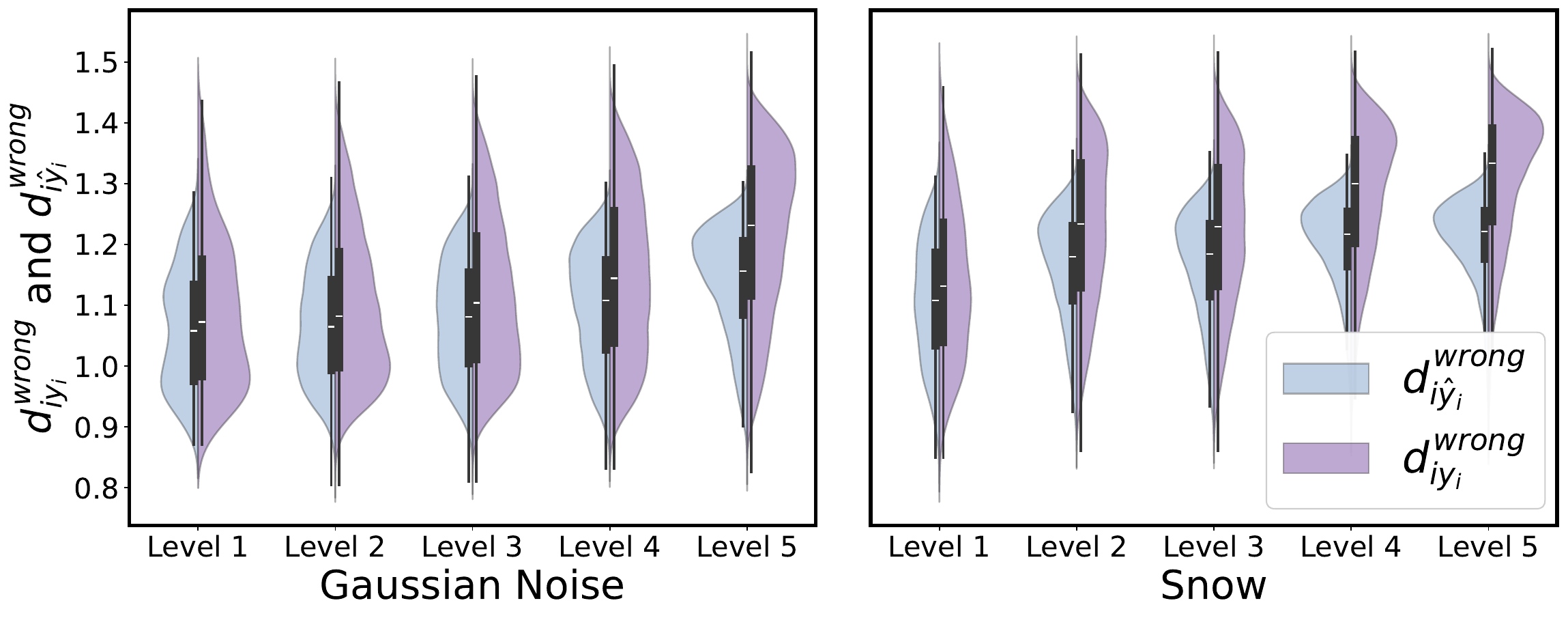}    
    \vspace{-10pt}
    \caption{\textbf{Violin plots of G-FCA and P-FCA distances for misclassified samples.} The plots show the distributions of distances from misclassified OOD samples to both G-FCA $d_{iy_i}^{\text{wrong}}$ and P-FCA $d_{i\hat{y}_i}^{\text{wrong}}$ under increasing Gaussian noise or Snow severity on ImageNet-C. The results reveal that misalignment becomes progressively more severe with higher corruption levels.}
    \vspace{-12pt}
    \label{fig:corruption-levels}
\end{figure}

NC3+ inspires us to understand performance degradation under domain shifts from the perspective of FCA distance. Prior work~\cite{ncooddetector} observed that feature embeddings of In-Distribution (ID) and OOD data tend to become orthogonal as training progresses. Building on this, we analyze the relationship between G-FCA distance and Pseudo-label FCA (P-FCA) distance to better understand the misalignment responsible for performance degradation. Figure~\ref{fig:histogram-tta} illustrates the distributions of G-FCA \(d_{iy_i}\) and P-FCA \(d_{i\hat{y}_i}\) distances of test samples $x_i \in D_{\text{test}}$ that are defined as:
\begin{align}
    \text{G-FCA:}\quad d_{iy_i} &= \left\| \frac{\mathbf{h}_i}{\|\mathbf{h}_i\|_2} - \frac{\mathbf{w}_{y_i}}{\|\mathbf{w}_{y_i}\|_2} \right\|_2, \\
    \text{P-FCA:}\quad d_{i\hat{y}_i} &= \left\| \frac{\mathbf{h}_i}{\|\mathbf{h}_i\|_2} - \frac{\mathbf{w}_{\hat{y}_i}}{\|\mathbf{w}_{\hat{y}_i}\|_2} \right\|_2.
\end{align}
where \( y_i \) is the ground-truth label and \( \hat{y}_i \) is the pseudo-label. Assume that the FCA distance \( d_{ij}^{(\cdot)} \) follows a normal distribution, \( d_{ij}^{(\cdot)} \sim N(\mu_{j}^{(\cdot)}, \sigma_{j}^{(\cdot)}) \), where \( \mu_j^{(\cdot)} \) and \( \sigma_j^{(\cdot)} \) represent the mean and variance. Here, $^{(\cdot)}$ indicates whether the test sample $x_i$ is classified correctly or incorrectly. For misclassified samples, \( \mu_{y_i}^{\text{wrong}} \) is significantly larger, reflecting severe misalignment with ground-truth class weights caused by domain shifts, while \( \mu_{\hat{y}_i}^{\text{wrong}} \) is smaller, indicating drift toward incorrect class weights. Figure~\ref{fig:corruption-levels} further quantifies this trend, showing that as corruption severity increases, the gap between \( \mu_{y_i}^{\text{wrong}} \) and \( \mu_{\hat{y}_i}^{\text{wrong}} \) widens. Figure~\ref{fig:histogram-tta} and Figure~\ref{fig:corruption-levels} reveal that feature-classifier misalignment is the primary driver of performance degradation for OOD data:
\begin{myfinding}{Sample-wise Misalignment in Adaptation} The feature embedding of an OOD sample tends to deviate from its corresponding classifier weight and drift toward that of an incorrect class, leading to misclassification. Formally, the means of G-FCA distance \(d_{iy_i}\) and P-FCA distance \(d_{i\hat{y}_i}\) satisfy:
\begin{equation}
    \mu_{\hat{y}_i}^{\text{correct}} = \mu_{y_i}^{\text{correct}} < \mu_{\hat{y}_i}^{\text{wrong}} \ll \mu_{y_i}^{\text{wrong}}.
\end{equation}
\end{myfinding}

NC3+ and the observation indicate that \textit{performance degradation under domain shifts stems from the displacement of feature embeddings away from corresponding classifier weights.} This misalignment increases $d_{iy_i}$ and reduces $d_{i\hat{y}_i}$, leading to erroneous predictions. 

\section{Method}
\label{sec:method}

\begin{figure*}[!h]
    \centering
    \includegraphics[width=0.85\textwidth]{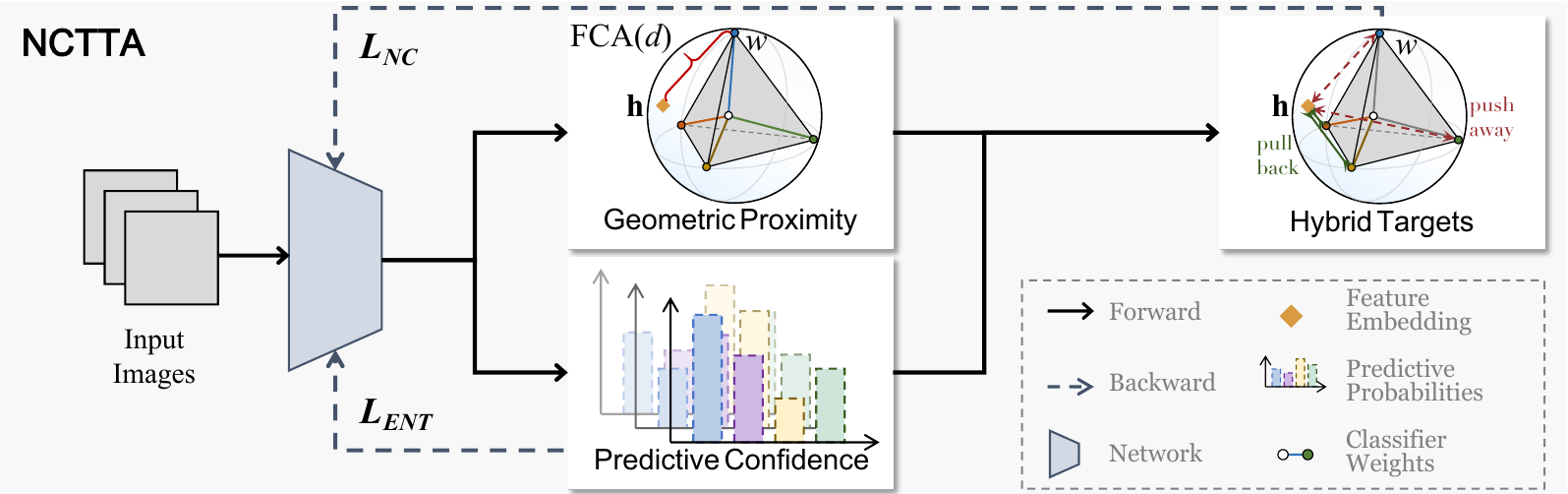}
    \vspace{-8pt}
    \caption{\textbf{Overview of our proposed NCTTA.} During test-time adaptation, NCTTA blends geometric proximity (FCA distance) and predictive confidence to form hybrid targets, pulling features toward plausible classifier weights while pushing away negatives via \(\mathcal{L}_{\text{NC}}\).}
  \vspace{-15pt}
    \label{fig:pipeline}
\end{figure*}

Building on the NC3+ property from Section~\ref{subsec:rnc-enc} and the observation from Section~\ref{subsec:tta-degradation}, we note that under domain shifts, OOD sample feature embeddings deviate from their corresponding classifier weights, thereby increasing G-FCA distance $d_{iy_i}$ and inducing misclassification as feature embeddings align with incorrect classifier weights. To address this issue, we propose a new method, NCTTA, that explicitly constrains $d_{iy_i}$ to realign feature embeddings with their correct classifier weights. Since TTA operates in an unlabeled regime, directly computing G-FCA distance $d_{iy_i}$ is infeasible. A straightforward workaround is to use the pseudo-label \(\hat{y}_i\) as a proxy for $y_i$ and constrain P-FCA distance $d_{i\hat{y}_i}$. However, as demonstrated in Section~\ref{subsec:tta-degradation}, pseudo-labels become unreliable under severe domain shifts. To overcome the limitation, NCTTA is designed as an NC-guided contrastive alignment mechanism with hybrid targets, blending geometric proximity and predictive confidence.

\paragraph{Neural Collapse in Test-Time Adaptation (NCTTA).}
During TTA, pretrained models are updated online using unlabeled data, which prevents direct computation of G-FCA distance $d_{iy_i}$. Section~\ref{subsec:tta-degradation} shows that the offset between $\mathbf{h}_i$ and $w_{y_i}$ makes P-FCA distance $\hat{y}_i$ unreliable. Therefore, simply constraining P-FCA distance $d_{i\hat{y}_i}$ is impractical. Our NCTTA replace $\hat{y}_i$ with dual-guided hybrid targets determined by \(\widetilde{\mathbf{y}}\in\mathbb{R}^{K}\) via:
\begin{equation}
    \widetilde{\mathbf{y}}_i = (1 - \alpha)\,\hat{d}_i + \alpha\, p_i,
\end{equation}
where $\hat{d}_i$ is obtained by first applying Z-score normalization to $d_i$, then taking the negative normalized distances followed by a softmax operation. 
This ensures that smaller FCA distances correspond to larger probability values. 
The coefficient $\alpha$ balances geometric proximity and predictive confidence. A larger $\widetilde{\mathbf{y}}_{ij}$ indicates that the feature embedding of \(x_i\) is geometrically closer to classifier weight $w_j$ and the model is more confident in class $j$. Therefore, we introduce the dual-guided order $\mathcal{O}_i=\text{argsort}(\widetilde{\mathbf{y}}_i)$, ranking classes in descending order of $\widetilde{\mathbf{y}}_i$. Obviously, we have $\mathcal{O}_i(\text{argmax}\ \widetilde{\mathbf{y}}_{i})=0$ and $\mathcal{O}_i(\text{argmin}\ \widetilde{\mathbf{y}}_{i})=K-1$. Hybrid targets are obtained using:

\begin{equation}
    \mathcal{T}_i=\{\mathcal{O}_i(0), \mathcal{O}_i(1), ..., \mathcal{O}_i(k - 1) \},
\end{equation}
where $\mathcal{T}_i$ is the set of the $k$ classes with the largest $\widetilde{\mathbf{y}}_i$ values. When $\alpha=1.0$ and $k=1$, $\mathcal{T}_i$ contains a single element, namely $\hat{y}_i$. Given $\mathcal{T}_i$, we then compute an NC-guided soft alignment loss:
\begin{equation}
    \mathcal{L}_{\text{NC}}(x_i)
= \ell(\{d_{ij}\}_{j\in \mathcal{T}_i},\{d_{ij}\}_{j\notin \mathcal{T}_i}),
\end{equation}
where $\ell$ enforces a monotonic preference for “closer to positives, farther from negatives.” Specifically, it is nonincreasing in $d_{ij}$ for $j\in \mathcal{T}_i$ which means $\partial\ell/\partial d_{ij}\geq 0\ (j\in \mathcal{T}_i)$ according to gradient descent~\cite{gradient_descent}, thereby encouraging reductions in $d_{ij}$ for positives; and it is nondecreasing in $d_{ij}$ for $j\notin \mathcal{T}_i$ which means $\partial\ell/\partial d_{ij}\leq 0\ (j\notin \mathcal{T}_i)$. We instantiate $\ell$ using InfoNCE-style, L2-style, or Triplet-style objectives, as detailed in Section~\ref{subsec:er-ablation}.

\paragraph{Final Objective.}
While $\mathcal{L}_{\text{NC}}$ leverages hybrid targets to mitigate misalignment between $\mathbf{h}_i$ and $w_{y_i}$, excessive noise in test samples can still destabilize adaptation. To enhance robustness, we integrate widely adopted techniques in the TTA literature into the final objective. We adopt an entropy-based filtering strategy inspired by prior works~\cite{EATA, DEYO, SAR}:
\begin{equation}
    S_{\text{ENT}} = \left\{ x_i \in D_{\text{test}} \ \middle| \ \mathcal{L}_{\text{ENT}}(x_i) < \gamma_{\text{ENT}} \right\},
\end{equation}
where $\gamma_{\text{ENT}}$ is a threshold filtering out high-entropy predictions, and $\mathcal{L}_{\text{ENT}}$ is the entropy minimization loss:
\begin{equation}
    \mathcal{L}_{\text{ENT}}(x_i) = -\sum_{j=0}^{K-1} p_j \log p_j.
\end{equation}
\begin{table*}[ht]
    \caption{
Classification accuracy (\%) on CIFAR-10-C and ImageNet-C under corruption severity level 5. The \textbf{bold} entries indicate the highest performance for each corruption type, and the \underline{underlined} entries indicate the next-highest performance.
}
    \vspace{-20pt}
    \label{tab:c&i}
    \begin{center}
    \begin{threeparttable}
    \LARGE
    \resizebox{1.0\linewidth}{!}{
    \begin{tabular}{l|ccc|cccc|cccc|cccc|c}
    
    \multicolumn{17}{c}{\textbf{CIFAR-10-C (ResNet50)}} \\
    \specialrule{1.2pt}{1pt}{1pt}
    Mild & Gauss. & Shot & Impul. & Defoc. & Glass & Motion & Zoom & Snow & Frost & Fog & Brit. & Contr. & Elastic & Pixel & JPEG & Avg. \\
    \midrule

    no\_adapt & 33.69 & 40.07 & 35.74 & 53.11 & 49.86 & 67.04 & 56.85 & 73.89 & 62.59 & 64.66 & \underline{89.00} & 43.57 & 74.37 & 41.96 & 74.43 & 57.39 \\
    BN\_adapt & 62.43$_{\pm 0.02}$ & 64.33$_{\pm 0.02}$ & 57.30$_{\pm 0.08}$ & 84.48$_{\pm 0.00}$ & 59.46$_{\pm 0.03}$ & 82.02$_{\pm 0.04}$ & 83.25$_{\pm 0.03}$ & 76.87$_{\pm 0.01}$ & 74.30$_{\pm 0.01}$ & 79.29$_{\pm 0.00}$ & 87.24$_{\pm 0.02}$ & 83.33$_{\pm 0.02}$ & 71.72$_{\pm 0.01}$ & 73.41$_{\pm 0.01}$ & 71.10$_{\pm 0.02}$ & 74.03 \\
    Tent & 64.71$_{\pm 0.03}$ & 66.97$_{\pm 0.03}$ & 59.61$_{\pm 0.01}$ & 84.97$_{\pm 0.03}$ & 60.96$_{\pm 0.04}$ & 82.48$_{\pm 0.04}$ & 83.93$_{\pm 0.01}$ & 77.78$_{\pm 0.01}$ & 75.42$_{\pm 0.00}$ & 80.00$_{\pm 0.04}$ & 87.78$_{\pm 0.01}$ & 83.54$_{\pm 0.00}$ & 72.62$_{\pm 0.05}$ & 74.73$_{\pm 0.01}$ & 72.32$_{\pm 0.03}$ & 75.19 \\
    EATA & 62.43$_{\pm 0.02}$ & 64.33$_{\pm 0.02}$ & 57.31$_{\pm 0.08}$ & 84.48$_{\pm 0.00}$ & 59.46$_{\pm 0.03}$ & 82.02$_{\pm 0.04}$ & 83.25$_{\pm 0.02}$ & 76.88$_{\pm 0.01}$ & 74.31$_{\pm 0.01}$ & 79.28$_{\pm 0.00}$ & 87.24$_{\pm 0.02}$ & 83.33$_{\pm 0.02}$ & 71.72$_{\pm 0.01}$ & 73.42$_{\pm 0.01}$ & 71.13$_{\pm 0.02}$ & 74.04 \\
    SAR & 63.74$_{\pm 0.07}$ & 65.87$_{\pm 0.06}$ & 58.47$_{\pm 0.02}$ & 84.70$_{\pm 0.01}$ & 60.51$_{\pm 0.02}$ & 82.22$_{\pm 0.02}$ & 83.62$_{\pm 0.02}$ & 77.26$_{\pm 0.03}$ & 74.74$_{\pm 0.02}$ & 79.74$_{\pm 0.02}$ & 87.40$_{\pm 0.01}$ & 83.29$_{\pm 0.02}$ & 72.24$_{\pm 0.03}$ & 74.15$_{\pm 0.00}$ & 71.87$_{\pm 0.02}$ & 74.67 \\
    NOTE & 55.17$_{\pm 0.03}$ & 60.13$_{\pm 0.15}$ & 55.13$_{\pm 0.06}$ & 74.53$_{\pm 0.00}$ & 59.43$_{\pm 0.08}$ & 78.61$_{\pm 0.04}$ & 75.72$_{\pm 0.08}$ & \underline{79.90$_{\pm 0.00}$} & \underline{76.46$_{\pm 0.01}$} & 78.18$_{\pm 0.05}$ & \textbf{90.91$_{\pm 0.01}$} & 70.43$_{\pm 0.04}$ & \underline{76.18$_{\pm 0.03}$} & 58.91$_{\pm 0.12}$ & 75.83$_{\pm 0.00}$ & 71.03 \\
    MEMO & 52.22$_{\pm 0.00}$ & 58.09$_{\pm 0.00}$ & 51.40$_{\pm 0.00}$ & 73.17$_{\pm 0.00}$ & 58.01$_{\pm 0.00}$ & 76.50$_{\pm 0.01}$ & 74.61$_{\pm 0.00}$ & 77.74$_{\pm 0.00}$ & 73.01$_{\pm 0.00}$ & 75.47$_{\pm 0.00}$ & 90.27$_{\pm 0.00}$ & 63.89$_{\pm 0.00}$ & \textbf{76.95$_{\pm 0.01}$} & 55.86$_{\pm 0.00}$ & 75.62$_{\pm 0.00}$ & 68.85 \\
    DeYO & \underline{67.61$_{\pm 0.27}$} & \underline{70.15$_{\pm 0.02}$} & \underline{63.15$_{\pm 0.15}$} & \underline{85.59$_{\pm 0.00}$} & \underline{62.98$_{\pm 0.03}$} & \underline{83.09$_{\pm 0.13}$} & \underline{84.57$_{\pm 0.00}$} & 79.05$_{\pm 0.00}$ & 76.35$_{\pm 0.02}$ & \underline{80.68$_{\pm 0.03}$} & 88.17$_{\pm 0.00}$ & \underline{83.57$_{\pm 0.00}$} & 74.16$_{\pm 0.04}$ & \underline{76.45$_{\pm 0.02}$} & \underline{74.13$_{\pm 0.08}$} & \underline{76.65} \\
    \rowcolor{cyan!10}NCTTA (ours) & \textbf{70.94$_{\pm 0.14}$} & \textbf{72.88$_{\pm 0.28}$} & \textbf{66.18$_{\pm 0.60}$} & \textbf{86.25$_{\pm 0.11}$} & \textbf{64.69$_{\pm 0.30}$} & \textbf{83.88$_{\pm 0.30}$} & \textbf{85.54$_{\pm 0.43}$} & \textbf{80.16$_{\pm 0.28}$} & \textbf{77.73$_{\pm 0.22}$} & \textbf{81.55$_{\pm 0.24}$} & 88.82$_{\pm 0.10}$ & \textbf{83.81$_{\pm 0.13}$} & 75.54$_{\pm 0.09}$ & \textbf{78.57$_{\pm 0.31}$} & \textbf{75.92$_{\pm 0.44}$} & \textbf{78.16} \\
    \specialrule{1.2pt}{1pt}{1pt}

    \multicolumn{17}{c}{\textbf{ImageNet-C (ViT-B/16)}} \\
    \specialrule{1.2pt}{1pt}{1pt}
    Mild & Gauss. & Shot & Impul. & Defoc. & Glass & Motion & Zoom & Snow & Frost & Fog & Brit. & Contr. & Elastic & Pixel & JPEG & Avg. \\
    \midrule

    no\_adapt & 34.93 & 33.67 & 36.41 & 32.38 & 22.22 & 36.61 & 31.10 & 22.15 & 26.73 & 53.16 & 61.17 & 50.27 & 31.96 & 54.52 & 55.87 & 38.88 \\
    Tent & 51.93$_{\pm 0.01}$ & 52.77$_{\pm 0.00}$ & 53.05$_{\pm 0.01}$ & 52.59$_{\pm 0.03}$ & 45.53$_{\pm 0.97}$ & 54.89$_{\pm 0.01}$ & 48.07$_{\pm 0.07}$ & 17.33$_{\pm 0.87}$ & 19.79$_{\pm 1.10}$ & 66.34$_{\pm 0.09}$ & 73.19$_{\pm 0.16}$ & 65.06$_{\pm 0.01}$ & 46.44$_{\pm 12.61}$ & 66.57$_{\pm 0.03}$ & 64.53$_{\pm 0.01}$ & 51.87 \\
    EATA & \underline{55.72$_{\pm 0.01}$} & \underline{56.94$_{\pm 0.00}$} & \underline{56.76$_{\pm 0.03}$} & 57.71$_{\pm 0.02}$ & \underline{55.81$_{\pm 0.05}$} & 62.07$_{\pm 0.00}$ & \underline{60.45$_{\pm 0.01}$} & 64.98$_{\pm 0.48}$ & 63.74$_{\pm 0.04}$ & 71.90$_{\pm 0.01}$ & 76.71$_{\pm 0.05}$ & \underline{67.81$_{\pm 0.24}$} & 66.29$_{\pm 0.00}$ & 72.48$_{\pm 0.01}$ & 69.25$_{\pm 0.00}$ & \underline{63.91} \\
    SAR & 51.93$_{\pm 0.01}$ & 52.90$_{\pm 0.01}$ & 53.06$_{\pm 0.00}$ & 52.66$_{\pm 0.00}$ & 47.08$_{\pm 0.09}$ & 54.88$_{\pm 0.05}$ & 48.42$_{\pm 0.01}$ & 26.60$_{\pm 10.31}$ & 34.87$_{\pm 4.86}$ & 66.28$_{\pm 0.13}$ & 72.79$_{\pm 0.01}$ & 64.67$_{\pm 0.00}$ & 50.80$_{\pm 0.58}$ & 66.40$_{\pm 0.00}$ & 64.36$_{\pm 0.00}$ & 53.97 \\
    NOTE & 35.11$_{\pm 0.00}$ & 33.86$_{\pm 0.00}$ & 36.58$_{\pm 0.00}$ & 32.71$_{\pm 0.00}$ & 22.39$_{\pm 0.00}$ & 36.92$_{\pm 0.00}$ & 31.30$_{\pm 0.00}$ & 22.33$_{\pm 0.00}$ & 26.90$_{\pm 0.00}$ & 53.14$_{\pm 0.00}$ & 62.25$_{\pm 2.02}$ & 51.18$_{\pm 0.00}$ & 32.04$_{\pm 0.00}$ & 54.64$_{\pm 0.00}$ & 55.95$_{\pm 0.00}$ & 39.15 \\
    MEMO & 43.78$_{\pm 0.00}$ & 43.25$_{\pm 0.01}$ & 45.83$_{\pm 0.00}$ & 32.90$_{\pm 0.03}$ & 28.46$_{\pm 0.03}$ & 45.67$_{\pm 0.07}$ & 38.25$_{\pm 0.12}$ & 30.08$_{\pm 0.06}$ & 34.63$_{\pm 0.15}$ & 54.80$_{\pm 0.20}$ & 69.99$_{\pm 0.37}$ & 56.43$_{\pm 0.16}$ & 34.47$_{\pm 0.01}$ & 62.56$_{\pm 0.03}$ & 59.61$_{\pm 0.04}$ & 45.38 \\
    DeYO & 55.68$_{\pm 0.01}$ & 56.92$_{\pm 0.02}$ & 56.42$_{\pm 0.03}$ & \underline{57.85$_{\pm 0.01}$} & 55.71$_{\pm 0.33}$ & \underline{62.68$_{\pm 0.10}$} & 46.53$_{\pm 0.38}$ & \underline{66.42$_{\pm 0.01}$} & \underline{65.46$_{\pm 0.01}$} & \underline{72.45$_{\pm 0.03}$} & \underline{78.38$_{\pm 0.01}$} & 66.52$_{\pm 0.05}$ & \underline{67.48$_{\pm 0.04}$} & \underline{73.63$_{\pm 0.00}$} & \underline{70.18$_{\pm 0.01}$} & 63.49 \\
    \rowcolor{cyan!10}NCTTA (ours) & \textbf{57.58$_{\pm 0.05}$} & \textbf{59.03$_{\pm 0.23}$} & \textbf{58.98$_{\pm 0.16}$} & \textbf{60.14$_{\pm 0.04}$} & \textbf{58.90$_{\pm 0.35}$} & \textbf{65.14$_{\pm 0.04}$} & \textbf{63.21$_{\pm 0.19}$} & \textbf{68.73$_{\pm 0.29}$} & \textbf{67.80$_{\pm 0.02}$} & \textbf{74.24$_{\pm 0.16}$} & \textbf{78.75$_{\pm 0.15}$} & \textbf{69.19$_{\pm 0.15}$} & \textbf{69.25$_{\pm 0.16}$} & \textbf{74.57$_{\pm 0.06}$} & \textbf{71.39$_{\pm 0.01}$} & \textbf{66.46} \\
    \specialrule{1.2pt}{1pt}{1pt}
    \end{tabular}
    }
    \end{threeparttable}
    \end{center}
    \vspace{-10pt}
\end{table*}
To jointly incorporate entropy and alignment confidence, we define a dynamic weight:
\begin{equation}
    \lambda_i = \frac{1}{\exp\left(\mathcal{L}_{\text{ENT}}(x_i) - \tau_{\text{ENT}}\right)} + \frac{\nu}{1 + \eta d_{i\hat{y}_i}},
\end{equation}
where $d_{i\hat{y}_i}$ denotes P-FCA distance, and $\tau_{\text{ENT}}$, $\nu$, $\eta$ control sensitivity to entropy and P-FCA. Combining entropy filtering, NC-guided soft alignment loss, and adaptive weighting, the final loss for each test sample is:
\begin{equation}
    \mathcal{L}_{\text{total}}(x_i) = \lambda_i \cdot \mathbb{I}_{x_i \in S_{\text{ENT}}} \cdot \left( \mathcal{L}_{\text{ENT}}(x_i) + \mathcal{L}_{\text{NC}}(x_i) \right),
\end{equation}
where $\mathbb{I}_{(\cdot)}$ is an indicator function that evaluates to 1 if $(\cdot)$ is true and 0 otherwise. This formulation jointly promotes confident predictions and robust alignment between feature embeddings and classifier weights, thereby improving robustness to domain shifts at test time.



\section{Experiments}
\subsection{Experimental Setup}

\begin{table}
    \centering
    \footnotesize
    \caption{Waterbirds average and worst-group accuracy (\%) on the target domain. \textbf{Bold} = best, \underline{underline} = second-best.}
    \vspace{-10pt}
    \renewcommand{\arraystretch}{0.95}
    \setlength{\tabcolsep}{3pt} 
    \resizebox{0.6\linewidth}{!}{
    \begin{tabular}{l|cc}
        \toprule
        \textbf{Method} & Avg. Acc & Worst-group Acc \\
        \midrule
        no\_adapt & 92.14 & 70.87 \\
        BN\_adapt & 91.20$_{\pm 0.00}$ & 61.78$_{\pm 0.02}$ \\
        Tent & 91.66$_{\pm 0.00}$ & 63.50$_{\pm 0.91}$ \\
        EATA & 91.53$_{\pm 0.01}$ & 63.14$_{\pm 0.34}$ \\
        SAR & 91.27$_{\pm 0.00}$ & 62.10$_{\pm 0.23}$ \\
        NOTE & 90.22$_{\pm 0.03}$ & 41.12$_{\pm 0.44}$ \\
        MEMO & 92.47$_{\pm 0.01}$ & 64.69$_{\pm 0.16}$ \\
        DeYO & \underline{92.90$_{\pm 0.07}$} & \underline{75.65$_{\pm 2.20}$} \\
        \rowcolor{cyan!10}NCTTA (ours) & \textbf{93.13$_{\pm 0.00}$} & \textbf{76.56$_{\pm 0.05}$} \\
        \bottomrule
    \end{tabular}
    }
    \label{tab:waterbirds}
    \vspace{-2mm}
\end{table}

\begin{figure}[htbp]
    \centering
    \captionsetup[subfigure]{justification=centering, labelformat=parens}
 
    \subcaptionbox{$\mathcal{L}_{\text{NC}}$}{%
        \includegraphics[width=0.320625\linewidth]{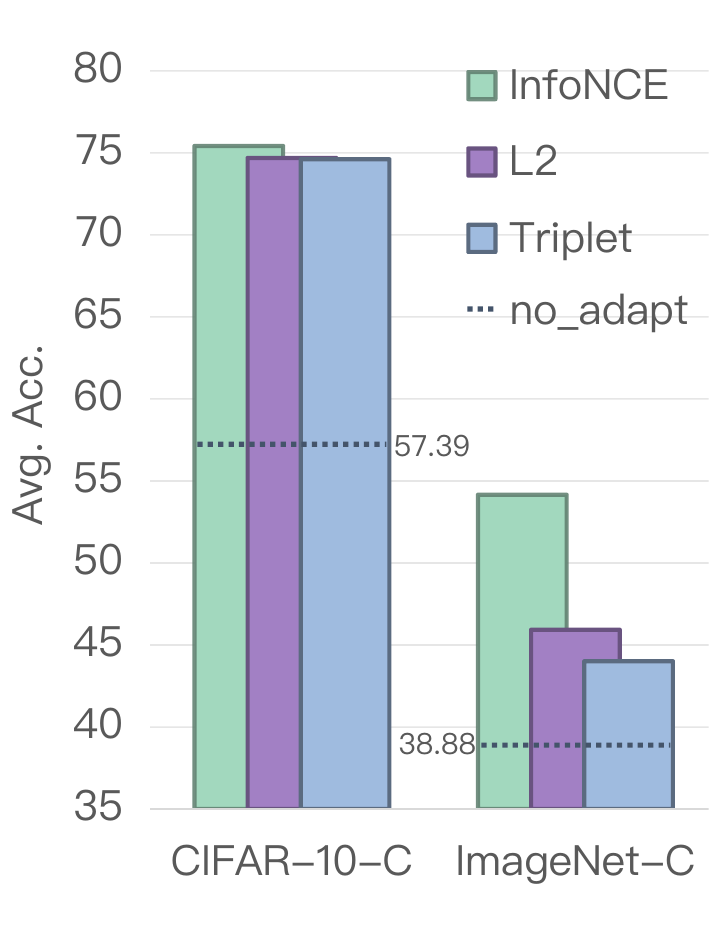}
    }
    \hfill
    \subcaptionbox{\(\alpha\) and \(k\)}{%
        \includegraphics[width=0.534375\linewidth]{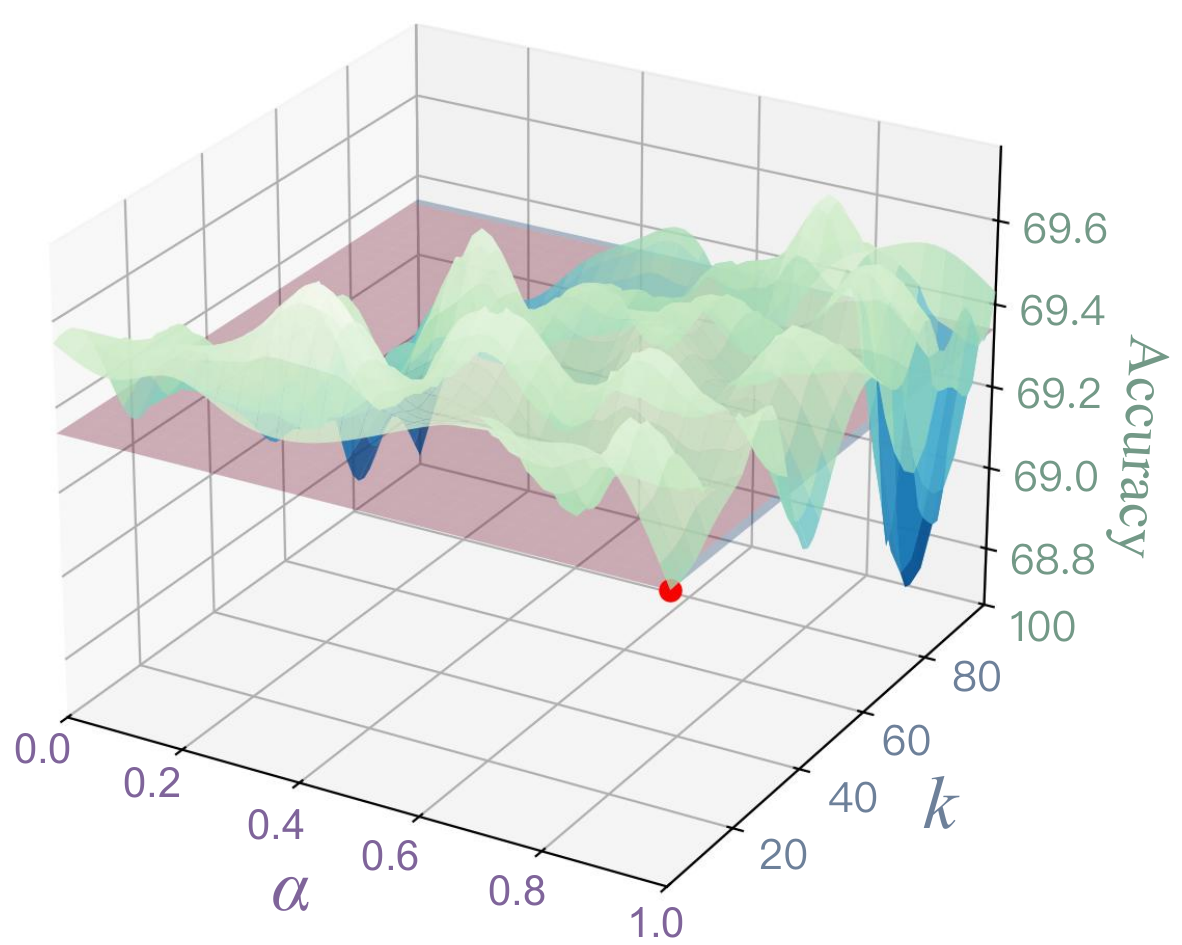}
    }
    \vspace{-8pt}
    \caption{\textbf{Ablation of \(\mathcal{L}_{\text{NC}}\), \(\alpha\) and \(k\).} (a) \(\mathcal{L}_{\text{NC}}\) is instantiated with three variants: InfoNCE-style, L2-style, and Triplet-style. (b) Sensitivity analysis of \(\alpha\) and \(k\), evaluated on ImageNet-C under the Contrast corruption at severity level 5.}
    \label{fig:ablation}
    \vspace{-12pt}
\end{figure}

The proposed method is evaluated on five widely used benchmarks: CIFAR-10/100-C~\cite{cifar10-100-C}, ImageNet-C~\cite{cifar10-100-C}, PACS~\cite{PACS} and Waterbirds~\cite{waterbirds}. Detailed descriptions of these datasets provided in Appendix B.1 and experiment results on CIFAR-100-C are shown in Appendix C. NCTTA is compared against a wide range of state-of-the-art TTA baselines, including no\_adapt (no adaptation), Tent~\cite{Tent}, BN\_adapt~\cite{bn_adapt}, NOTE~\cite{note}, MEMO~\cite{Memo}, SAR~\cite{SAR}, EATA~\cite{EATA} and DeYO~\cite{DEYO}. Furthermore, we provide the results of EATA-C~\cite{tan2025uncertainty}, SAR$^2$~\cite{niu2025adapt}, COME~\cite{zhang2024come}, and AdaDEM~\cite{ma2025decoupled} on CIFAR-10/100-C and ImageNet-C in the Appendix for completeness. Detailed descriptions of these baselines, along with their hyperparameter configurations, are provided in Appendix B.2. The backbone used in the experiments are ResNet50~\cite{resnet} for CIFAR-10/100-C, WaterBirds and PACS, and ViT-B/16~\cite{vit} for ImageNet-C. Experiments are conducted on the TTAB~\cite{TTAB} benchmark, with algorithmic parameters set to their default values unless otherwise specified.

\begin{table*}[ht]
    \vspace{-10pt} 
    \caption{Classification accuracy (\%) on PACS. The \textbf{bold} entries indicate the highest performance for each corruption type, and the \underline{underlined} entries indicate the next-highest performance.}
    \label{tab:pacs}
    \vspace{-10pt}
    \centering
    \resizebox{0.8\textwidth}{!}{%
    \begin{tabular}{l|ccc|ccc|ccc|c}
        \toprule
        Source &
        \multicolumn{3}{c|}{\textbf{Art}} &
        \multicolumn{3}{c|}{\textbf{Photo}} &
        \multicolumn{3}{c|}{\textbf{Cartoon}} &
        \textbf{Avg.} \\
        Target & Cartoon & Photo & Sketch & Cartoon & Art & Sketch & Art & Photo & Sketch & \\
        \midrule
        no\_adapt & 65.61 & 97.66 & 57.32 & 39.55 & 73.24 & 43.93 & 75.63 & 90.48 & 72.23 & 68.41 \\

        BN\_adapt & 74.67$_{\pm 0.03}$ & 96.85$_{\pm 0.02}$ & 69.32$_{\pm 0.16}$ & 64.39$_{\pm 0.01}$ & 77.17$_{\pm 0.11}$ & 45.85$_{\pm 0.09}$ & 82.137$_{\pm 0.03}$ & 94.69$_{\pm 0.14}$ & 73.05$_{\pm 0.09}$ & 75.35 \\

        Tent & 75.17$_{\pm 0.02}$ & 96.93$_{\pm 0.02}$ & 71.52$_{\pm 0.08}$ & 65.09$_{\pm 0.11}$ & 77.39$_{\pm 0.21}$ & \underline{48.80$_{\pm 0.07}$} & 83.11$_{\pm 0.00}$ & \underline{94.89$_{\pm 0.06}$} & 75.00$_{\pm 0.08}$ & 76.43 \\

        EATA & 74.70$_{\pm 0.03}$ & 96.85$_{\pm 0.02}$ & 69.01$_{\pm 0.02}$ & 64.32$_{\pm 0.00}$ & 77.22$_{\pm 0.14}$ & 45.58$_{\pm 0.00}$ & 82.13$_{\pm 0.03}$ & 94.69$_{\pm 0.14}$ & 73.05$_{\pm 0.05}$ & 75.28 \\

        SAR & 74.94$_{\pm 0.03}$ & 96.85$_{\pm 0.02}$ & 69.38$_{\pm 0.05}$ & 64.51$_{\pm 0.03}$ & 77.31$_{\pm 0.22}$ & 45.85$_{\pm 0.08}$ & 82.37$_{\pm 0.03}$ & 94.81$_{\pm 0.10}$ & 73.63$_{\pm 0.12}$ & 75.52 \\

        NOTE & 70.72$_{\pm 0.16}$ & 97.68$_{\pm 0.00}$ & 65.47$_{\pm 0.14}$ & 47.95$_{\pm 0.06}$ & 75.13$_{\pm 0.07}$ & \textbf{50.23$_{\pm 0.14}$} & 79.70$_{\pm 0.01}$ & 93.33$_{\pm 0.06}$ & \textbf{76.17$_{\pm 0.07}$} & 72.93 \\

        MEMO & 69.51$_{\pm 0.01}$ & \underline{98.00$_{\pm 0.00}$} & 61.37$_{\pm 0.01}$ & 45.42$_{\pm 0.02}$ & 74.67$_{\pm 0.02}$ & 44.28$_{\pm 0.00}$ & 80.16$_{\pm 0.02}$ & 93.87$_{\pm 0.01}$ & 74.44$_{\pm 0.00}$ & 71.30 \\

        DeYO & \underline{76.82$_{\pm 0.27}$} & 96.95$_{\pm 0.00}$ & \underline{70.75$_{\pm 0.10}$} & \underline{66.77$_{\pm 0.00}$} & \underline{77.74$_{\pm 0.00}$} & 46.31$_{\pm 0.15}$ & \underline{84.30$_{\pm 0.11}$} & \textbf{95.07$_{\pm 0.08}$} & \underline{75.29$_{\pm 0.09}$} & \underline{76.67} \\

        \rowcolor{cyan!10}
            NCTTA (ours) & \textbf{77.21$_{\pm 0.99}$} & \textbf{98.00$_{\pm 0.69}$} & \textbf{71.68$_{\pm 0.00}$} & \textbf{70.94$_{\pm 0.00}$} & \textbf{77.78$_{\pm 0.28}$} & 46.17$_{\pm 0.00}$ & \textbf{84.80$_{\pm 0.00}$} & 94.41$_{\pm 0.35}$ & 75.26$_{\pm 0.00}$ & \textbf{77.36} \\
        \bottomrule
    \end{tabular}
    }
    \vspace{-10pt} 
\end{table*}

We designed our experiments to answer the following questions: 
\textbf{RQ1:} How does NCTTA perform in comparison to baseline methods across various scenarios, including biased and wild scenarios that resemble real-world situations?
\textbf{RQ2:} How to select the specific form of \(\mathcal{L}_{\text{NC}}\)? And is the proposed NCTTA effective? 
\textbf{RQ3:} What role does the NCTTA play? 
\subsection{Main Results}
\label{subsec:er-b}
To thoroughly evaluate the effectiveness of NCTTA, three challenging experimental setups are designed: 1) Mild Scenario: Each test batch consists of samples from a single corruption type or domain, simulating stable domain shifts. 2) Small Batch Size: The model processes one test sample per iteration, testing the adaptability of TTA methods under extreme resource constraints. 3) Continual Test-Time Adaptation (CTTA): The test data undergoes continual shifts across multiple domains, presenting a dynamic and evolving distribution. These setups are designed to assess NCTTA's robustness and generalization capabilities under diverse and realistic conditions. The concrete implementation of NCTTA will be detailed in Section~\ref{subsec:er-ablation}. We adopt Eq.~\ref{eq:infonce} as the default instantiation of \(\mathcal{L}_\text{NC}\).

In the mild shift scenario, where test samples are corrupted but drawn from a single domain (Table~\ref{tab:c&i}, Table~\ref{tab:pacs} and Table~\ref{tab:waterbirds}), NCTTA consistently outperforms prior methods across all benchmarks. On CIFAR-10-C and ImageNet-C, NCTTA achieves the highest average accuracy, ranking first on nearly all corruption types. Similar gains are observed on PACS and Waterbirds, where NCTTA handles spurious correlations and domain shifts more effectively than existing methods. These results demonstrate the benefit of our feature-classifier alignment under domain shifts. Moreover, under certain corruptions, such as Snow and Frost on ImageNet-C, Tent performs even worse than no\_adapt. In contrast, NCTTA can mitigate the shortcomings of the $\mathcal{L}_{\text{ENT}}$ loss simply by incorporating $\mathcal{L}_{\text{NC}}$, without the need for additional augmentations or carefully designs.

In the wild shift scenario, where test domains change continually (Table~\ref{tab:ctta}) or the batch size is restricted to 1 (Table~\ref{tab:ib1}), NCTTA demonstrates robust and consistent performance, showcasing its ability to handle challenging real-world conditions. Specifically, in the CTTA setting, NCTTA achieves an impressive accuracy of 71.32\%, significantly surpassing all other methods, which underscores its effectiveness in adapting to continuously evolving test domains. Furthermore, in the batch size 1 setting, a scenario that limits the availability of test samples and poses a severe challenge for most methods, NCTTA attains an accuracy of 59.55\%, outperforming Tent by more than 10.36\%. This substantial margin highlights the superiority of NCTTA in scenarios with limited batch sizes, where efficient and reliable adaptation is critical. Overall, these findings affirm the stability, adaptability, and scalability of NCTTA across diverse and dynamically shifting test-time conditions, making it a robust solution for real-world deployment.

\begin{table}[htbp]
  \centering
  \vspace{-5pt}
  \caption{Ablation study of the proposed components. The table reports average classification accuracy (\%) across the 15 corruption types of CIFAR-10-C and ImageNet-C at severity level 5. \textbf{Bold} numbers denote the best performance.}
  \label{tab:ablation-ours}
  \vspace{-10pt}
  \resizebox{0.99\linewidth}{!}{%
    \begin{tabular}{lcccccc}
      \toprule
      & $\mathcal{L}_{\text{ENT}}$ & $\mathcal{L}_{\text{NC}}$ & $S_{\text{ENT}}$ & $\lambda$ & CIFAR-10-C & ImageNet-C \\
      \midrule
      (1)Tent  & \ding{51} &                 &           &           & 75.19 & 51.13 \\
      (2)SAR   & \ding{51} &                 & \ding{51} &           & 74.81 & 52.27 \\
      (3)      &           & \ding{51}       &           &           & 75.31 & 53.17 \\
      (4)      &           & \ding{51}       & \ding{51} &           & 74.95 & 51.39 \\
      (5)      & \ding{51} & \ding{51}       &           &           & 76.25 & 54.84 \\
      (6)      & \ding{51} & \ding{51}       & \ding{51} &           & 75.64 & 57.34 \\
      (7)      &           & \ding{51}       & \ding{51} & \ding{51} & 76.87 & 53.97 \\
      (8)      & \ding{51} &                 & \ding{51} & \ding{51} & 76.55 & 53.20 \\
      \rowcolor{cyan!10}
      (9)NCTTA(ours)    & \ding{51} & \ding{51}       & \ding{51} & \ding{51} & \textbf{78.10} & \textbf{66.50} \\
      \bottomrule
    \end{tabular}%
  }
  \vspace{-10pt}
\end{table}

\subsection{Ablation Studies}
\label{subsec:er-ablation}
In this work, we instantiate three variants of \(\mathcal{L}_\text{NC}\), InfoNCE-style, L2-style, and Triplet-style, defined as:
\begin{equation}
\label{eq:infonce}
\mathcal{L}_\text{NC}^{\text{InfoNCE}}(x_i)
=
-\log \frac{\frac{1}{|\mathcal{T}|}\sum_{j\in\mathcal{T}}\exp\!\left(\langle \frac{\mathbf{h}_i}{\|\mathbf{h}_i\|_2}, \frac{w_{j}}{\|w_{j}\|_2} \rangle  \right)}
{\sum_{j=0}^{K-1} \exp\!\left(\langle \frac{\mathbf{h}_i}{\|\mathbf{h}_i\|_2}, \frac{w_j}{\|w_j\|_2} \rangle  \right)},
\end{equation}
\begin{equation}
\begin{aligned}
\mathcal{L}_\text{NC}^{\text{L2}}(x_i)
& =
\frac{1}{|\mathcal{T}|}\sum_{j\in\mathcal{T}}\left \|\frac{\mathbf{h}_i}{\|\mathbf{h}_i\|_2}-\frac{w_{j}}{\|w_{j}\|_2}\right \|_2\\
& -\frac{1}{K-|\mathcal{T}|}\sum_{j\notin\mathcal{T}}\left \|\frac{\mathbf{h}_i}{\|\mathbf{h}_i\|_2}-\frac{w_{j}}{\|w_{j}\|_2}\right \|_2,
\end{aligned}
\end{equation}
\begin{equation}
\begin{aligned}
\mathcal{L}_\text{NC}^{\text{Triplet}}(x_i)
&= \max\bigl(0,\,
\frac{1}{|\mathcal{T}|}\sum_{j\in\mathcal{T}}\left \|\frac{\mathbf{h}_i}{\|\mathbf{h}_i\|_2}-\frac{w_{j}}{\|w_{j}\|_2}\right \|_2 \\
&\quad - \min_{j\notin\mathcal{T}}\left \|\frac{\mathbf{h}_i}{\|\mathbf{h}_i\|_2}-\frac{w_{j}}{\|w_{j}\|_2}\right \|_2 + \tau_{\text{margin}}\bigr),
\end{aligned}
\end{equation}
where \(\langle \cdot, \cdot \rangle\) denotes the similarity measure, instantiated as cosine similarity in our experiments. \(\tau_{\text{margin}}\) is the margin hyperparameter, which we set to \(1.0\). We evaluate the three formulations on the 15 corruption types of CIFAR-10-C and ImageNet-C, setting \(k=1\) for each. As shown in Figure~\ref{fig:ablation}(a), all variants consistently improve performance by alleviating sample-wise misalignment, underscoring the importance of realigning feature embeddings and the corresponding classifier weights. Among them, the InfoNCE-style loss delivers the most stable and substantial gains. Therefore, we adopt it as the default instantiation of \(\mathcal{L}_\text{NC}\) in all subsequent experiments.

\begin{table}[ht]
    \vspace{-5pt} 
    \caption{Classification accuracy (\%) on ImageNet-C under the CTTA scenario across severity levels 1 to 5. The \textbf{bold} entries indicate the highest performance for each corruption type, and the \underline{underlined} entries indicate the next-highest performance.}
    \label{tab:ctta}
    \vspace{-10pt}
    \centering
    \footnotesize
    \resizebox{0.99\linewidth}{!}{\begin{tabular}{l|ccccc|c}
        \toprule
        CTTA & Level 1 & Level 2 & Level 3 & Level 4 & Level 5 & Avg. \\
        \midrule
        no\_adapt & 70.78 & 63.65 & 58.85 & 50.44 & 38.82 & 56.51 \\

        Tent & 76.01 & 71.47 & 68.42 & 62.27 & 53.17 & 66.27 \\

        EATA & 77.85 & 74.13 & 71.35 & 66.20 & \underline{58.65} & 69.64 \\

        SAR & 75.68 & 71.09 & 67.90 & 61.91 & 53.53 & 66.02 \\

        NOTE & 63.82 & 63.68 & 58.83 & 50.61 & 39.18 & 55.22 \\

        MEMO & 74.41 & 69.32 & 65.05 & 56.93 & 45.55 & 62.25 \\

        DeYO & \underline{78.40} & \underline{74.49} & \underline{71.75} & \underline{66.28} & 57.88 & \underline{69.76} \\

        \rowcolor{cyan!10}NCTTA (ours) & \textbf{78.82} & \textbf{75.44} & \textbf{72.84} & \textbf{68.15} & \textbf{61.11} & \textbf{71.27} \\
        \bottomrule
    \end{tabular}}
    \vspace{-10pt}
\end{table}

\begin{table*}[ht]
    \caption{Classification accuracy (\%) on ImageNet-C under the batch size = 1 scenario across severity level 5. The \textbf{bold} entries indicate the highest performance for each corruption type, and the \underline{underlined} entries indicate the next-highest performance.}
    \label{tab:ib1}
    \vspace{-20pt}
    \begin{center}
    \begin{threeparttable}
    \LARGE
    \resizebox{1.0\linewidth}{!}{
    \begin{tabular}{l|ccc|cccc|cccc|cccc|c}
    \specialrule{1.2pt}{1pt}{1pt}
    \textbf{batch size = 1} & Gauss. & Shot & Impul. & Defoc. & Glass & Motion & Zoom & Snow & Frost & Fog & Brit. & Contr. & Elastic & Pixel & JPEG & Avg. \\
    \midrule

    no\_adapt & 34.93 & 33.67 & 36.41 & 32.38 & 22.22 & 36.61 & 31.10 & 22.15 & 26.73 & 53.16 & 61.17 & 50.27 & 31.96 & 54.52 & 55.87 & 38.88 \\

    Tent & 47.15$_{\pm 0.67}$ & 46.40$_{\pm 1.03}$ & 47.51$_{\pm 0.02}$ & 45.19$_{\pm 1.04}$ & 34.01$_{\pm 0.36}$ & 47.57$_{\pm 0.33}$ & 39.52$_{\pm 0.22}$ & 37.29$_{\pm 4.59}$ & 40.03$_{\pm 2.24}$ & 58.97$_{\pm 0.05}$ & 69.43$_{\pm 0.65}$ & 60.61$_{\pm 0.06}$ & 39.88$_{\pm 0.19}$ & 62.76$_{\pm 2.00}$ & 61.49$_{\pm 0.23}$ & 49.19 \\

    EATA & 48.56$_{\pm 0.16}$ & 48.17$_{\pm 0.28}$ & 49.62$_{\pm 0.09}$ & 48.17$_{\pm 0.80}$ & 39.10$_{\pm 0.84}$ & 51.34$_{\pm 0.38}$ & 44.37$_{\pm 1.44}$ & 45.01$_{\pm 2.01}$ & 46.00$_{\pm 3.08}$ & 64.84$_{\pm 0.11}$ & 72.59$_{\pm 0.52}$ & 61.45$_{\pm 0.43}$ & 47.65$_{\pm 3.65}$ & 63.97$_{\pm 1.56}$ & 64.03$_{\pm 0.12}$ & 52.99 \\

    SAR & 40.32$_{\pm 0.04}$ & 39.79$_{\pm 0.41}$ & 42.77$_{\pm 1.08}$ & 38.73$_{\pm 0.08}$ & 27.73$_{\pm 0.08}$ & 45.13$_{\pm 0.21}$ & 36.85$_{\pm 0.36}$ & 30.86$_{\pm 3.35}$ & 36.04$_{\pm 0.93}$ & 58.06$_{\pm 0.14}$ & 68.43$_{\pm 0.42}$ & 56.47$_{\pm 0.44}$ & 37.16$_{\pm 1.27}$ & 63.34$_{\pm 19.33}$ & 60.95$_{\pm 1.43}$ & 45.51 \\

    NOTE & 35.25$_{\pm 0.16}$ & 33.56$_{\pm 0.06}$ & 36.13$_{\pm 0.88}$ & 32.53$_{\pm 0.20}$ & 22.42$_{\pm 0.25}$ & 36.82$_{\pm 0.10}$ & 30.91$_{\pm 0.49}$ & 22.47$_{\pm 0.21}$ & 27.08$_{\pm 0.45}$ & 53.43$_{\pm 0.02}$ & 61.41$_{\pm 0.29}$ & 50.15$_{\pm 0.33}$ & 31.87$_{\pm 0.45}$ & 54.39$_{\pm 0.10}$ & 55.83$_{\pm 1.21}$ & 38.95 \\

    MEMO & 37.31$_{\pm 0.34}$ & 35.19$_{\pm 0.40}$ & 39.04$_{\pm 0.01}$ & 32.13$_{\pm 0.27}$ & 22.57$_{\pm 0.40}$ & 38.44$_{\pm 0.16}$ & 31.81$_{\pm 0.00}$ & 23.78$_{\pm 0.26}$ & 28.87$_{\pm 0.00}$ & 52.91$_{\pm 0.02}$ & 62.27$_{\pm 0.06}$ & 52.30$_{\pm 0.06}$ & 31.96$_{\pm 0.23}$ & 54.77$_{\pm 0.06}$ & 56.27$_{\pm 0.28}$ & 39.97 \\

    DeYO & \underline{50.72$_{\pm 0.48}$} & \underline{50.58$_{\pm 0.13}$} & \underline{52.45$_{\pm 1.54}$} & \underline{51.47$_{\pm 0.43}$} & \underline{42.75$_{\pm 7.59}$} & \underline{54.10$_{\pm 0.07}$} & \underline{45.91$_{\pm 0.25}$} & \underline{52.73$_{\pm 0.28}$} & \underline{51.84$_{\pm 1.90}$} & \underline{66.36$_{\pm 0.16}$} & \underline{73.10$_{\pm 0.09}$} & \underline{62.42$_{\pm 0.50}$} & \underline{52.11$_{\pm 1.54}$} & \underline{66.14$_{\pm 0.93}$} & \underline{64.12$_{\pm 0.58}$} & \underline{55.79} \\

    \rowcolor{cyan!10}NCTTA (ours) & \textbf{52.32$_{\pm 0.18}$} & \textbf{53.03$_{\pm 0.33}$} & \textbf{55.01$_{\pm 0.06}$} & \textbf{53.09$_{\pm 0.24}$} & \textbf{48.53$_{\pm 0.91}$} & \textbf{57.50$_{\pm 0.97}$} & \textbf{53.35$_{\pm 0.16}$} & \textbf{59.64$_{\pm 0.17}$} & \textbf{56.79$_{\pm 0.12}$} & \textbf{69.83$_{\pm 0.71}$} & \textbf{75.59$_{\pm 0.61}$} & \textbf{65.09$_{\pm 0.38}$} & \textbf{57.55$_{\pm 0.23}$} & \textbf{70.34$_{\pm 0.55}$} & \textbf{67.70$_{\pm 0.23}$} & \textbf{59.69} \\
    \specialrule{1.2pt}{1pt}{1pt} 
    \end{tabular}
    }
    \end{threeparttable}
    \end{center}
    \vspace{-20pt}
\end{table*}

To assess the efficacy of the proposed hybrid targets and the sensitivity of NCTTA to their associated parameters \(\alpha\) and \(k\), we conduct a controlled study summarized in Figure~\ref{fig:ablation}(b). Properly tuning \(\alpha\) and \(k\) yields superior performance compared to the degenerate setting that solely constrains the P-FCA distance (i.e., \(\alpha=1.0\), \(k=1\)). This result indicates that hybrid targets effectively mitigate the unreliability of pseudo-labels caused by sample-wise misalignment in adaptation.

Table~\ref{tab:ablation-ours} reports ablations on CIFAR-10-C and ImageNet-C at severity level 5. Introducing \(\mathcal{L}_\text{NC}\) substantially improves robustness and complements the widely used entropy regularization \(\mathcal{L}_\text{ENT}\). The full NCTTA configuration outperforms all partial variants, highlighting the synergistic effect of each component in strengthening feature-classifier alignment and increasing resilience to severe corruptions.

\subsection{Role of NCTTA}

\label{subsec:role}
\begin{figure}[htbp]
    \vspace{-10pt}
    \centering
    \includegraphics[width=0.85\linewidth]{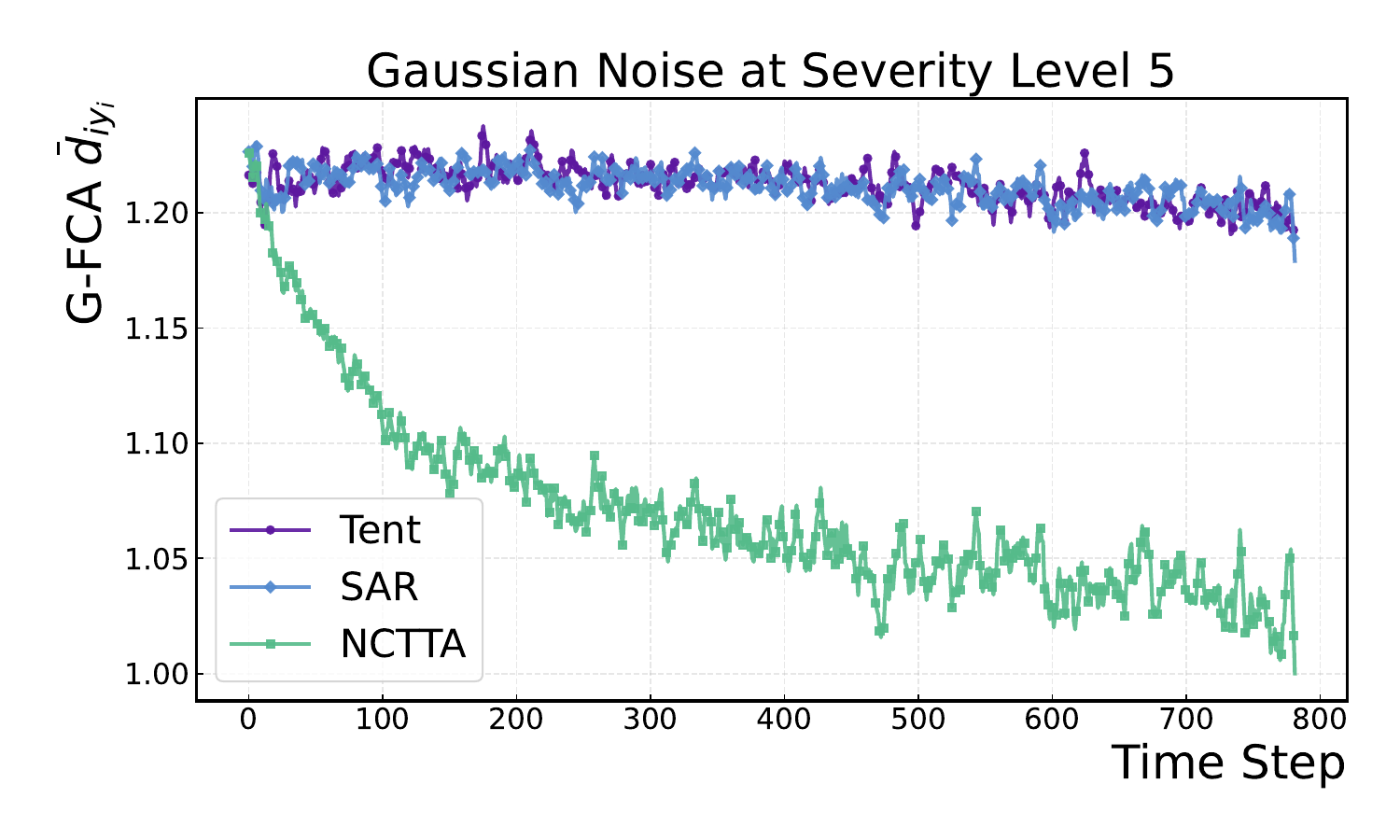}
    \vspace{-15pt}
    \caption{\textbf{Comparison of G-FCA distance under Gaussian noise (ImageNet-C, ViT-B/16).} The plot compares $d_{iy_i}$ for Tent, SAR and NCTTA on ImageNet-C under Gaussian noise with severity level 5. The results show that NCTTA consistently achieves lower $d_{y_i}$ compared to Tent and SAR, demonstrating better feature-classifier alignment and enhanced robustness.}
    \label{fig:visualdis}
    \vspace{-15pt}
\end{figure}

To empirically validate the effectiveness of NCTTA, we systematically computed the average G-FCA distance \( \bar{d}_{iy_i} \) for each TTA batch and compared the results against baseline methods SAR and Tent. As illustrated in Figure~\ref{fig:visualdis}, NCTTA consistently achieves lower \( \bar{d}_{iy_i} \) values compared to the baseline approaches across both CIFAR-10-C and ImageNet-C under Gaussian noise with severity level 5. This indicates that NCTTA effectively mitigates feature-classifier misalignment, thereby enhancing the model's robustness to severe domain shifts.

\begin{figure}[htbp]
    \centering
    \captionsetup[subfigure]{justification=centering, labelformat=parens}
    
    \subcaptionbox{Tent}{%
        \includegraphics[width=0.48\linewidth]{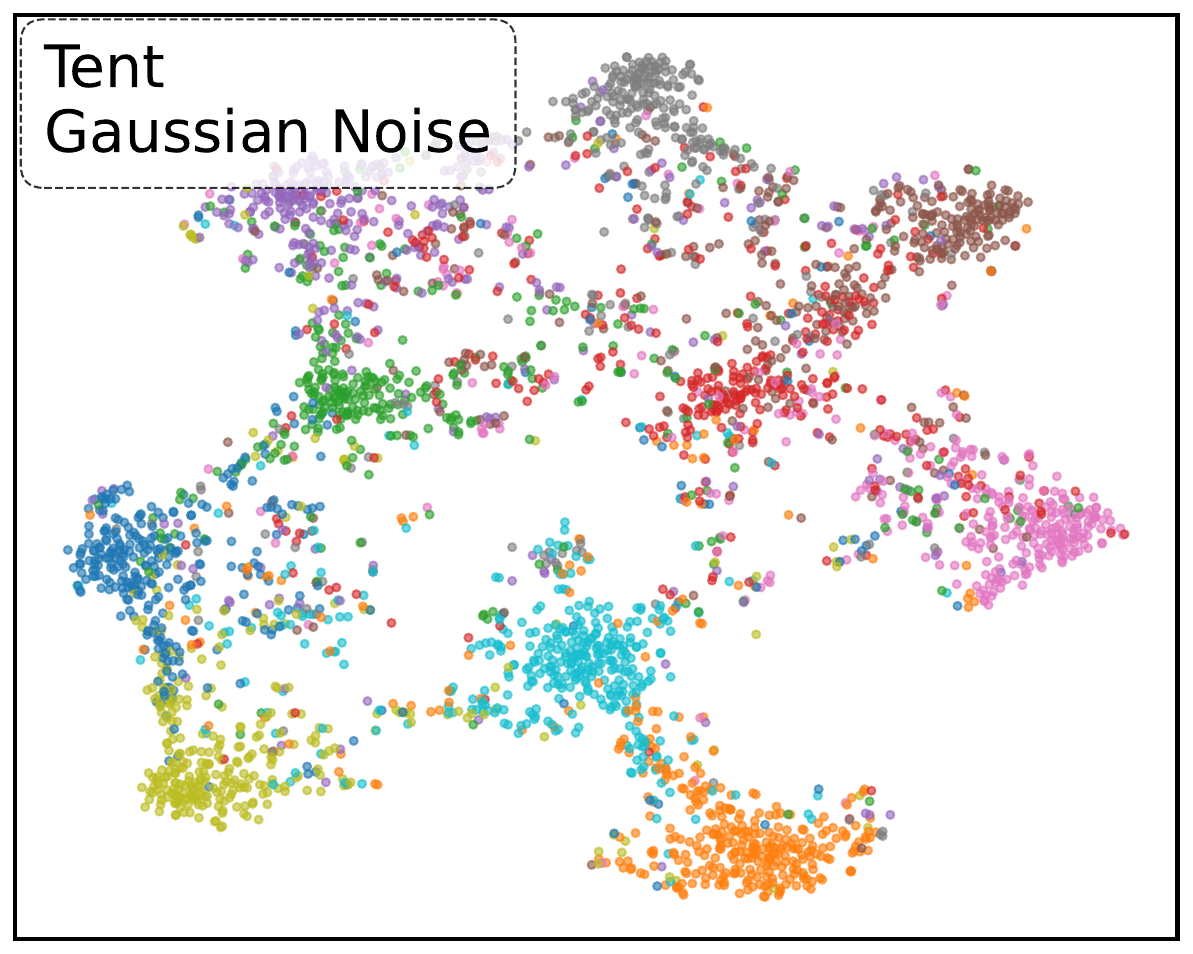}
    }
    \hfill
    \subcaptionbox{NCTTA}{%
        \includegraphics[width=0.48\linewidth]{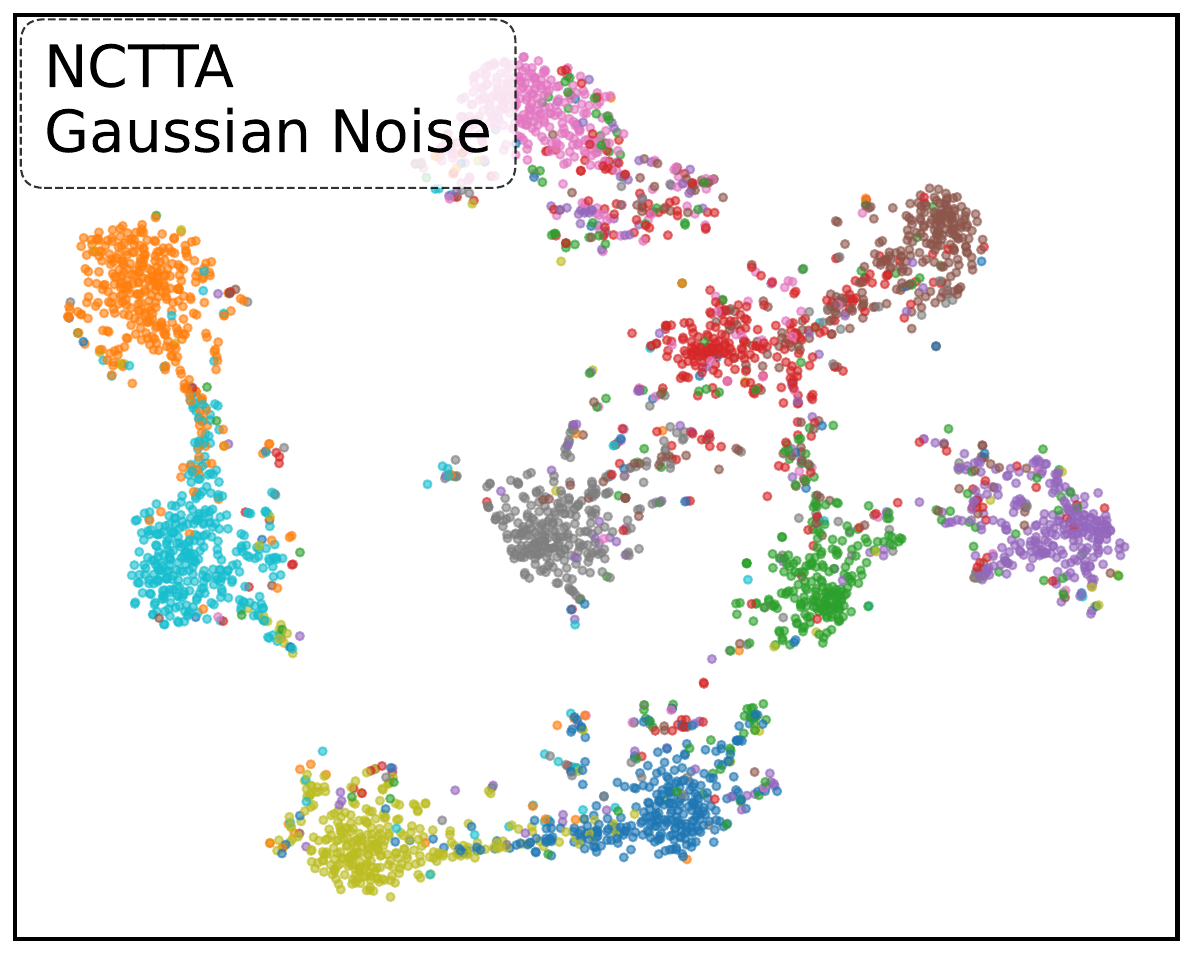}
    }
    \vspace{-10pt}
    \caption{\textbf{t-SNE comparison under Gaussian noise.} The figure compares t-SNE visualizations of feature representations for Tent and NCTTA on CIFAR-10-C under Gaussian noise with severity level 5. NCTTA forms more distinct and well-separated clusters compared to Tent, enhancing the model's discriminative power under severe corruption.}
    \label{fig:tsne}
    \vspace{-12pt}
\end{figure}

Furthermore, t-SNE visualizations~\cite{van2008visualizing} presented in Figure~\ref{fig:tsne} substantiate these quantitative findings. The feature embeddings processed with NCTTA form more distinct and well-separated clusters compared to those obtained using Tent. This enhanced clustering demonstrates NCTTA's superior capability in maintaining discriminative feature spaces, which is crucial for accurate classification in OOD scenarios. Collectively, these results underscore that NCTTA not only improves feature-classifier alignment but also effectively distinguishes features under various distributional shifts, reinforcing the method’s robustness.

\section{Conclusion and Limitation}
\label{sec:conclusion}
In this work, we bridge NC theory and TTA by extending NC3 to the sample-wise regime, formulating NC3+. We theoretically and empirically show that, during the TPT, each sample’s feature embedding converges toward its corresponding classifier weight, and we reveal that domain shifts primarily harm performance by inducing sample-wise misalignment. Building on these insights, we propose NCTTA, an NC-guided adaptation framework that realigns feature embeddings to classifier weights using hybrid targets blending geometric proximity (FCA distance) and predictive confidence. 

NCTTA consistently improves robustness across CIFAR-10-C/100-C, ImageNet-C, PACS, and Waterbirds, including challenging settings such as CTTA and batch size $=1$, demonstrating strong generality and stability. Notably, NCTTA yielded significant improvements in the G-FCA distance \(d_{iy_i}\), alongside enhanced feature clustering, as shown in t-SNE visualizations. Our work leaves several aspects unexplored. Specifically, further research is required to harness the potential of NC properties for computing or learning class prototypes in mini-batch TTA settings. Exploring the applicability of NC in vision-language models (VLMs)~\cite{radford2021learning, tpt, tda, boostadapter, huang2025cosmic, chen2025test} also remains an open direction for future investigation. This work provides a new perspective for advancing both the theoretical understanding and practical development of TTA methods.

\section*{Acknowledgment}
This work was supported by the National Natural Science Foundation of China (Grant No. 92467204 and 62472249), the Shenzhen Science and Technology Program (Grant No. JCYJ20220818101014030, KJZD20240903102300001, and JCYJ20250604145014018), and the Natural Science Foundation of Top Talent of SZTU (Grant No. GDRC202413). We thank the anonymous reviewers for their efforts, helping improve the quality of this paper.

{
    \small
    \bibliographystyle{ieeenat_fullname}
    \bibliography{main}
}

\end{document}